\documentclass[11pt]{article}

\usepackage[hyperref]{acl}
\usepackage{times}
\usepackage{latexsym}
\usepackage[T1]{fontenc}
\usepackage[utf8]{inputenc}
\usepackage{microtype}
\usepackage{inconsolata}

\usepackage{graphicx}
\usepackage{amsmath}
\usepackage{amssymb}
\usepackage{booktabs}
\usepackage{multirow}
\usepackage{array}
\usepackage{subfigure}
\usepackage{algorithm}
\usepackage{algorithmic}
\usepackage{bm}
\usepackage{xcolor}
\usepackage{enumitem}

\title{Multi-View Deformable Convolution Meets Visual Mamba for Coronary Artery Segmentation}

\author{
  \textbf{Xiaochan Yuan}$^{1}$ \and 
  \textbf{Pai Zeng}$^{1}$ \\
  $^{1}$Sichuan Agricultural University \\
  \small $^{1}$Email: \texttt{202306653@stu.sicau.edu.cn}, \texttt{202006728@stu.sicau.edu.cn}
}

\begin{document}

\maketitle

% ============================================================================
% ABSTRACT
% ============================================================================
\begin{abstract}
Accurate segmentation of coronary arteries from computed tomography angiography (CTA) images is of paramount clinical importance for the diagnosis and treatment planning of cardiovascular diseases. However, coronary artery segmentation remains challenging due to the inherent multi-branching and slender tubular morphology of the vasculature, compounded by severe class imbalance between foreground vessels and background tissue. Conventional convolutional neural network (CNN)-based approaches struggle to capture long-range dependencies among spatially distant vascular structures, while Vision Transformer (ViT)-based methods incur prohibitive computational overhead that hinders deployment in resource-constrained clinical settings. Motivated by the recent success of state space models (SSMs) in efficiently modeling long-range sequential dependencies with linear complexity, we propose MDSVM-UNet, a novel two-stage coronary artery segmentation framework that synergistically integrates multidirectional snake convolution (MDSConv) with residual visual Mamba (RVM). In the encoding stage, we introduce MDSConv, a deformable convolution module that learns adaptive offsets along three orthogonal anatomical planes---sagittal, coronal, and axial---thereby enabling comprehensive multi-view feature fusion that faithfully captures the elongated and tortuous geometry of coronary vessels. In the decoding stage, we design an RVM-based upsampling decoder block that leverages selective state space mechanisms to model inter-slice long-range dependencies while preserving linear computational complexity. Furthermore, we propose a progressive two-stage segmentation strategy: the first stage performs coarse whole-image segmentation to guide intelligent block extraction, while the second stage conducts fine-grained block-level segmentation to recover vascular details and suppress false positives. Extensive experiments on the large-scale ImageCAS benchmark demonstrate that MDSVM-UNet achieves substantial improvements over the state-of-the-art ImageCAS baseline, with gains of 5.41\% in Dice Similarity Coefficient, 8.5456 in Hausdorff Distance, and 0.8093 in Average Hausdorff Distance, while maintaining a competitive parameter count of 26.7M.
\end{abstract}

% ============================================================================
% 1. INTRODUCTION
% ============================================================================
\section{Introduction}
\label{sec:introduction}

Cardiovascular disease (CVD) constitutes the leading cause of mortality among noncommunicable diseases worldwide, accounting for approximately 17.9 million deaths annually and representing a critical global health challenge \cite{li2018retrieval}. Among the spectrum of cardiovascular pathologies, coronary artery disease (CAD) stands out as a predominant contributor to morbidity and mortality, with coronary artery stenosis identified as a principal risk factor that can precipitate myocardial infarction, heart failure, and sudden cardiac death \cite{gharleghi2022survey}. In clinical practice, computed tomography angiography (CTA) has emerged as a widely adopted noninvasive imaging modality for the diagnosis and treatment planning of coronary artery disease, owing to its capacity to provide high-resolution three-dimensional (3D) visualization of the coronary vasculature \cite{zeng2023imagecas}. The diagnostic workflow typically involves radiologists manually delineating coronary artery boundaries, followed by extraction and quantification of stenotic segments to inform therapeutic decisions. Nevertheless, this manual annotation process is exceedingly time-consuming, error-prone, and subject to substantial inter-observer variability, rendering it increasingly impractical given the exponential growth in medical imaging volumes \cite{li2018retrieval, lou2024review}. Consequently, developing automated and accurate coronary artery segmentation methods has become an urgent clinical imperative.

Over the past decade, significant research efforts have been devoted to vascular segmentation, encompassing both traditional machine learning approaches and deep learning-based methods. Early approaches relied on handcrafted feature representations and rule-based techniques, including region growing methods \cite{chen2013coronary}, active contour models \cite{shang2013vessel}, and tracking algorithms \cite{zhou2010coronary}. While these methods demonstrated initial promise, they are inherently limited by their dependence on domain-specific expert knowledge and suffer from poor scalability and generalization. The advent of deep learning has fundamentally transformed the landscape of medical image segmentation, with U-Net \cite{ronneberger2015unet} and its numerous architectural variants establishing themselves as the \textit{de facto} standard for dense prediction tasks. Building upon the foundational encoder-decoder paradigm, researchers have proposed a series of innovations aimed at preserving fine-grained spatial details for precise boundary delineation, notably UNet++ \cite{zhou2018unetpp}, which introduces nested dense skip connections to bridge the semantic gap between encoder and decoder feature maps. To effectively capture cross-slice contextual information in volumetric medical images, 3D extensions of these architectures have been developed, including 3D U-Net \cite{cicek2016_3dunet}, nnU-Net \cite{isensee2021nnunet}, and CAS-Net \cite{dong2023casnet}. More recently, Transformer-based architectures \cite{hatamizadeh2022unetr, cao2022swinunet, xu2023octa, lou2024review} have garnered substantial attention in vascular segmentation due to their inherent ability to capture global long-range dependencies through self-attention mechanisms.

Despite the notable progress achieved by these U-shaped architectures, several fundamental challenges persist in coronary artery segmentation that remain inadequately addressed \cite{gharleghi2022survey, qi2023dsconv}. \textbf{First}, coronary arteries exhibit a distinctive morphology characterized by slender, elongated, and highly branched tubular structures, yet standard convolution operations employ regular grid sampling patterns that fail to incorporate geometric and topological priors inherent to vascular anatomy, thereby limiting their capacity to faithfully extract tubular features. \textbf{Second}, existing U-shaped networks are inherently constrained in their ability to model long-range inter-slice dependencies that are critical for maintaining vascular continuity across adjacent CT slices, while Transformer-based alternatives impose quadratic computational complexity with respect to input sequence length, rendering them impractical for high-resolution 3D medical imaging. \textbf{Third}, conventional skip connections in encoder-decoder architectures typically perform naive feature concatenation that indiscriminately introduces irrelevant background noise, exacerbating the already severe foreground-background imbalance and making it difficult to distinguish coronary arteries from surrounding anatomical structures.

To address these intertwined challenges, we propose MDSVM-UNet, a two-stage coronary artery segmentation framework that leverages multidirectional snake convolution and residual visual Mamba. Recent advances in state space models (SSMs), exemplified by Mamba \cite{gu2023mamba}, have demonstrated remarkable efficiency in modeling long-range sequential dependencies while maintaining linear computational complexity, positioning them as compelling alternatives to both CNNs and Vision Transformers for processing high-resolution volumetric data. Inspired by these advances, our framework introduces the following key innovations:

\begin{enumerate}[leftmargin=*,itemsep=2pt]
    \item \textbf{Multidirectional Snake Convolution (MDSConv):} We propose a novel deformable convolution module that extends dynamic snake convolution \cite{qi2023dsconv} to three orthogonal anatomical planes---sagittal, coronal, and axial. By learning adaptive deformation offsets along each anatomical axis and fusing multi-view features, MDSConv enables the model to comprehensively capture the elongated and tortuous geometry of coronary vessels from complementary perspectives.
    
    \item \textbf{Residual Visual Mamba (RVM) Decoder:} We design an upsampling decoder block that integrates residual visual Mamba layers with dense spatial pooling for multi-scale context generation. This architecture effectively models long-range inter-slice dependencies while maintaining linear computational complexity, substantially reducing parameter overhead compared to Transformer-based decoders.
    
    \item \textbf{Two-Stage Progressive Segmentation:} We propose a coarse-to-fine segmentation strategy in which the first stage performs whole-image segmentation on downsampled CTA volumes to guide intelligent block extraction, and the second stage conducts refined block-level segmentation to recover fine vascular details, reduce false positives, and improve overall continuity and smoothness of the segmentation results.
\end{enumerate}

Extensive experiments on the large-scale ImageCAS benchmark \cite{zeng2023imagecas} demonstrate that our proposed two-stage MDSVM-UNet achieves state-of-the-art performance, outperforming the previous best method by 5.41\% in Dice Similarity Coefficient (DSC), 8.5456 in Hausdorff Distance (HD), and 0.8093 in Average Hausdorff Distance (AHD), while maintaining a competitive model size of 26.7M parameters.

% ============================================================================
% 2. RELATED WORK
% ============================================================================
\section{Related Work}
\label{sec:related_work}

\subsection{Coronary Artery Segmentation}
\label{subsec:coronary_seg}

Vascular segmentation, and coronary artery segmentation in particular, occupies a central role in the broader domain of medical image analysis~\cite{cao2022swinunet,schaap2009evaluation,es2023ragasautomatedevaluationretrieval}. With the rapid proliferation of deep learning techniques ~\cite{tang2022few, tang2022optimal, MeasurLink} in medical imaging, learning-based coronary artery segmentation has emerged as the dominant research paradigm, progressively superseding traditional rule-based approaches. Lei et al.~\cite{lei2020coronary} introduced a 3D attention-enhanced fully convolutional network (FCN) \cite{shelhamer2017fcn} for automated coronary artery segmentation, integrating attention mechanisms into the FCN architecture to selectively emphasize semantically informative features extracted from computed tomography coronary angiography (CTCA) images. While their approach demonstrated the feasibility of end-to-end segmentation from downscaled input volumes, the interpolation operations inherent in image resizing inevitably incur pixel-level information loss, compromising the fidelity of fine vascular structures.

Huang et al.~\cite{huang2018coronary} adopted an alternative strategy of partitioning original CTCA images into smaller volumetric patches and feeding them into a 3D U-Net for coronary lumen segmentation. However, the use of small input patches constrains the effective receptive field, potentially causing the network to overlook global anatomical context and local structural details simultaneously. Kong et al.~\cite{kong2020tree} proposed a tree-structured convolutional gated recurrent unit (TS ConvGRU) specifically designed to learn the hierarchical tree-like anatomical structure of the coronary artery system. Their framework integrates TS ConvGRU into a two-stage pipeline, employing 3D U-Net for initial coarse segmentation, followed by centerline extraction and subsequent fine segmentation with TS ConvGRU layers. Wolterink et al.~\cite{wolterink2019graph} explored graph convolutional networks (GCN) for coronary artery segmentation, predicting spatial positions of vertices on tubular surface meshes to automatically extract surface representations encompassing the entire coronary tree. The incorporation of mesh-based geometric priors ~\cite{wang2023improving,wang2024enhancing,wang2025evaluating, wang2025fine} improved segmentation overlap and accuracy without requiring explicit mesh-level supervision.

More recently, Dong et al.~\cite{dong2023casnet} proposed CAS-Net, a multi-attention, multi-scale 3D deep network that designs attention-guided feature fusion modules, scale-aware feature modules for dynamic receptive field adjustment, and multi-scale feature aggregation modules for enriched semantic representations. Zeng et al.~\cite{zeng2023imagecas} released ImageCAS, the first large-scale public dataset comprising 1,000 CTA images with high-quality annotations, alongside a strong two-stage baseline combining multi-scale patch fusion with progressive refinement. Post-processing-based approaches~\cite{wang2024cardcros, wang2025vision, wang2025pargo, wang2025wilddoc, tang2022youcan,tang2024mtvqa,tang2024textsquare,wu2022twostage} have also gained traction: Qiu et al.~\cite{qiu2023corsegrec} proposed CorSegRec, which integrates image segmentation with centerline reconnection and geometric reconstruction to preserve topological connectivity, while Wang et al.~\cite{wang2024cardcros} introduced a data-driven refinement method for repairing disconnected arterial structures. These post-processing methods are complementary to our approach and can be applied downstream to further enhance segmentation quality.

\subsection{Deformable Convolution}
\label{subsec:deformable_conv}

Standard convolution operations employ rigid, regular grid sampling patterns that impose geometric constraints on feature extraction, fundamentally limiting their ability to adapt to objects with irregular or non-rigid morphologies. Yu and Koltun~\cite{yu2016dilated} proposed dilated (atrous) convolutions to systematically aggregate multi-scale contextual information by increasing the effective receptive field without sacrificing spatial resolution, thereby replacing conventional upsampling and downsampling operations and reducing information loss. Dai et al.~\cite{dai2017deformable} identified the geometric rigidity of standard CNN building blocks as an inherent limitation and introduced deformable convolutions and deformable region-of-interest (ROI) pooling~\cite{fei2025advancing,feng2023unidoc,feng2024docpedia,feng2025dolphin,fu2024ocrbench}, which augment standard modules with learnable spatial offsets that enable free-form sampling in the neighborhood of each spatial location without requiring additional supervision.

While deformable convolutions exhibit improved adaptability to geometric deformations of diverse objects, they remain suboptimal for the precise extraction of elongated, tortuous, and densely branched tubular structures characteristic of coronary vasculature. Recognizing this gap, Qi et al.~\cite{qi2023dsconv} proposed dynamic snake convolution (DSConv), a topology-aware deformable convolution method specifically tailored for tubular structure segmentation. DSConv introduces deformable offsets into the convolution kernel and employs an iterative offset propagation strategy that sequentially selects observation targets for each processing position, ensuring attention continuity along the tubular centerline while preventing excessive receptive field dispersion due to large offsets. Our proposed MDSConv extends this concept to three orthogonal anatomical planes, enabling comprehensive multi-directional feature fusion.

\subsection{State Space Models}
\label{subsec:ssm}

State space models (SSMs), exemplified by the recently proposed Mamba architecture \cite{gu2023mamba}, have emerged as formidable competitors to both CNNs and Vision Transformers (ViTs) across a variety of sequence modeling tasks. The high spatial resolution inherent in 3D medical imaging volumes poses significant computational challenges for ViT-based methods, whose quadratic self-attention complexity severely degrades both throughput and memory efficiency. Mamba addresses these limitations through a selective mechanism and hardware-aware algorithmic design that enables efficient modeling of long-range dependencies with linear computational complexity, thereby substantially improving both training and inference efficiency.

The success of Mamba has catalyzed extensive exploration of its application in computer vision and medical image analysis. Liu et al.~\cite{liu2024a_swinumamba} proposed Swin-UMamba, a Mamba-based model specifically designed for medical image segmentation that leverages ImageNet-pretrained weights to enhance representation learning. Liu et al.~\cite{liu2024b_vmamba} introduced VMamba (Vision Mamba), incorporating bidirectional SSM blocks for data-dependent global visual context modeling and positional embeddings for position-aware visual understanding. Liao et al.~\cite{liao2024lightmunet} proposed LightM-UNet, a pure Mamba-based architecture that employs residual visual Mamba layers to extract deep semantic features and model long-range spatial dependencies with linear complexity, achieving competitive performance with significantly fewer parameters. Wu et al.~\cite{wu2024ultralight} introduced parallel visual Mamba blocks that partition feature channels into four parallel residual visual Mamba streams, optimizing long-range spatial information acquisition while reducing computational load. Zhang et al.~\cite{zhang2024vmunetv2} proposed VM-UNet-V2, which introduces vision state space (VSS) blocks for comprehensive contextual information capture and semantic-detail fusion for enhanced multi-level feature integration.

However, the aforementioned methods ~\cite{li2024real,lu2024bounding,lu2025prolonged,niu2025cme,guo2025seed1,liu2023spts,zhao2024harmonizing,zhao2024multi,zhao2025tabpedia,sun2025attentive,yu2016dilated,yu2025benchmarking,li2026dtp,cui2026diffusion,zhu2026textpecker} primarily target general-purpose medical image segmentation tasks and do not explicitly account for the distinctive elongated tubular morphology of coronary arteries. In contrast, our work represents \textit{the first application of visual Mamba to coronary artery segmentation}, combining structure-aware deformable convolutions with efficient long-range dependency modeling to address the unique challenges of this clinically important task.

% ============================================================================
% 3. METHOD
% ============================================================================
\section{Methodology}
\label{sec:method}

\begin{figure*}[t]
    \centering
    \includegraphics[width=\textwidth]{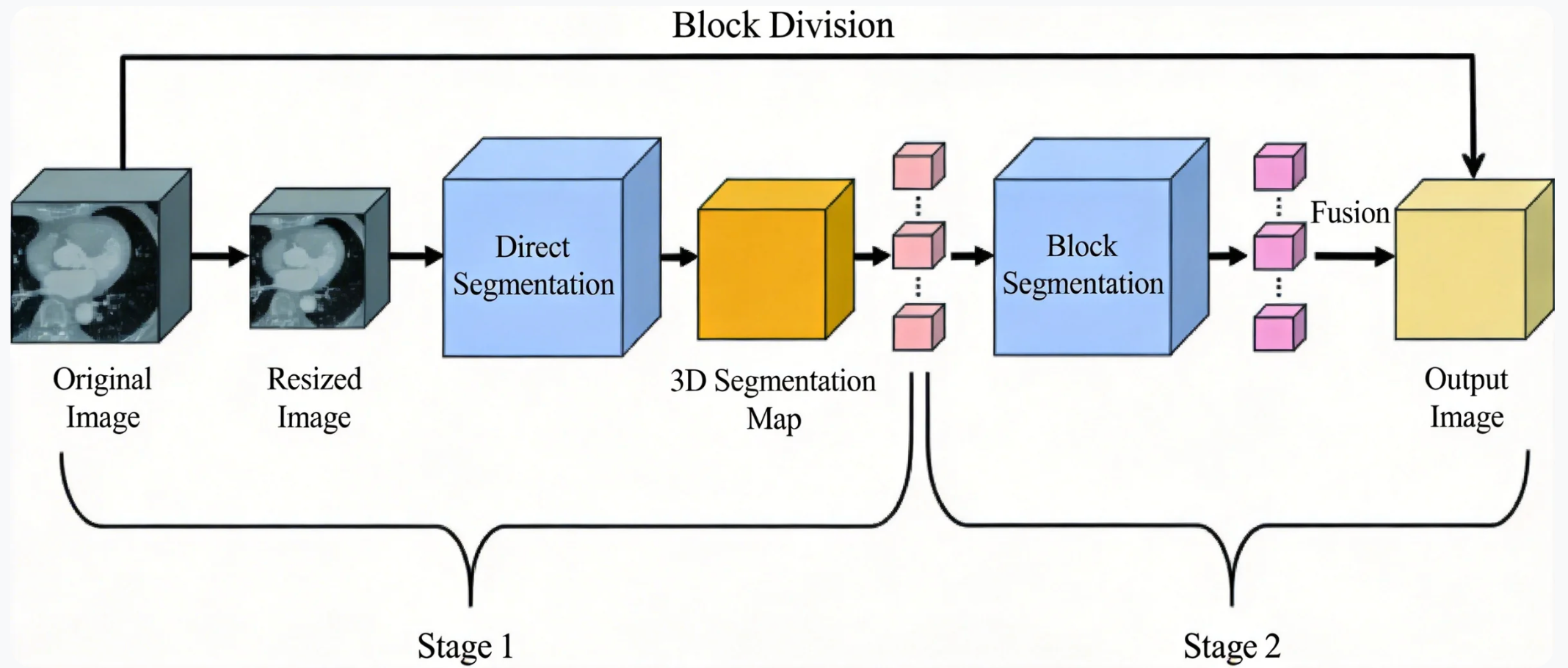}
    \caption{Overview of the proposed two-stage coronary artery segmentation framework. In Stage~1, the input CTA volume is downsampled and segmented using MDSVM-UNet to produce a coarse segmentation map. The coarse results guide the extraction of $64 \times 64 \times 64$ voxel blocks from the original resolution. In Stage~2, MDSVM-UNet performs fine-grained block-level segmentation, and the results from all blocks are merged to produce the final output.}
    \label{fig:framework}
\end{figure*}

We present MDSVM-UNet, a two-stage coronary artery segmentation framework that integrates multidirectional snake convolution (MDSConv) in the encoder and residual visual Mamba (RVM) in the decoder, organized within a progressive coarse-to-fine segmentation pipeline. An overview of the complete framework is illustrated in Figure~\ref{fig:framework}, and the detailed network architecture is shown in Figure~\ref{fig:network}.

\subsection{Two-Stage Segmentation Pipeline}
\label{subsec:pipeline}

Existing CTA-based coronary artery segmentation methods can be broadly categorized into two paradigms: direct segmentation and block-based segmentation. Direct segmentation methods process entire CTA volumes using a single neural network, benefiting from large receptive fields and producing probability maps that require minimal post-processing. However, they necessitate downsampling the input to lower resolutions due to GPU memory constraints, inevitably sacrificing fine-grained vascular details. Block-based segmentation methods partition the volumetric data into smaller 3D patches (e.g., $64 \times 64 \times 64$ voxels) for processing, which preserves spatial resolution but constrains the receptive field and increases susceptibility to false positive predictions.

To leverage the complementary strengths of both paradigms while mitigating their respective limitations, we propose a progressive two-stage segmentation pipeline:

\textbf{Stage 1: Coarse Whole-Image Segmentation.} In the first stage, the original CTA volume is downsampled to a resolution of $128 \times 128 \times 64$ and fed into MDSVM-UNet for coarse whole-image segmentation. The resulting coarse segmentation map captures the global anatomical structure and spatial distribution of the coronary artery tree, providing essential guidance for the subsequent block extraction process. Specifically, after performing inference on the entire training, validation, and test sets, the coarse segmentation results are used to identify and extract coronary artery-containing regions from the original full-resolution images.

\textbf{Stage 2: Fine-Grained Block Segmentation.} The coarse segmentation results from Stage~1 guide the extraction of $64 \times 64 \times 64$ voxel blocks from the original high-resolution CTA volumes, ensuring that the extracted blocks contain a high concentration of coronary artery information. These blocks are then re-input into MDSVM-UNet for fine-grained segmentation, enabling the network to learn local vascular details within a smaller spatial neighborhood. Finally, the segmentation results from all blocks are merged to produce the complete segmentation output.

This progressive coarse-to-fine strategy effectively resolves the inherent tension between global context modeling and local detail preservation. Compared to existing two-stage dynamic segmentation approaches \cite{wu2022twostage}, our method exclusively utilizes the first-stage results as guidance for block extraction rather than directly incorporating them into the final output, thereby substantially reducing false positive artifacts introduced by inaccurate coarse segmentation while simultaneously improving the continuity and smoothness of the segmented coronary arteries.

\subsection{Network Architecture}
\label{subsec:architecture}

\begin{figure*}[t]
    \centering
    \includegraphics[width=\textwidth]{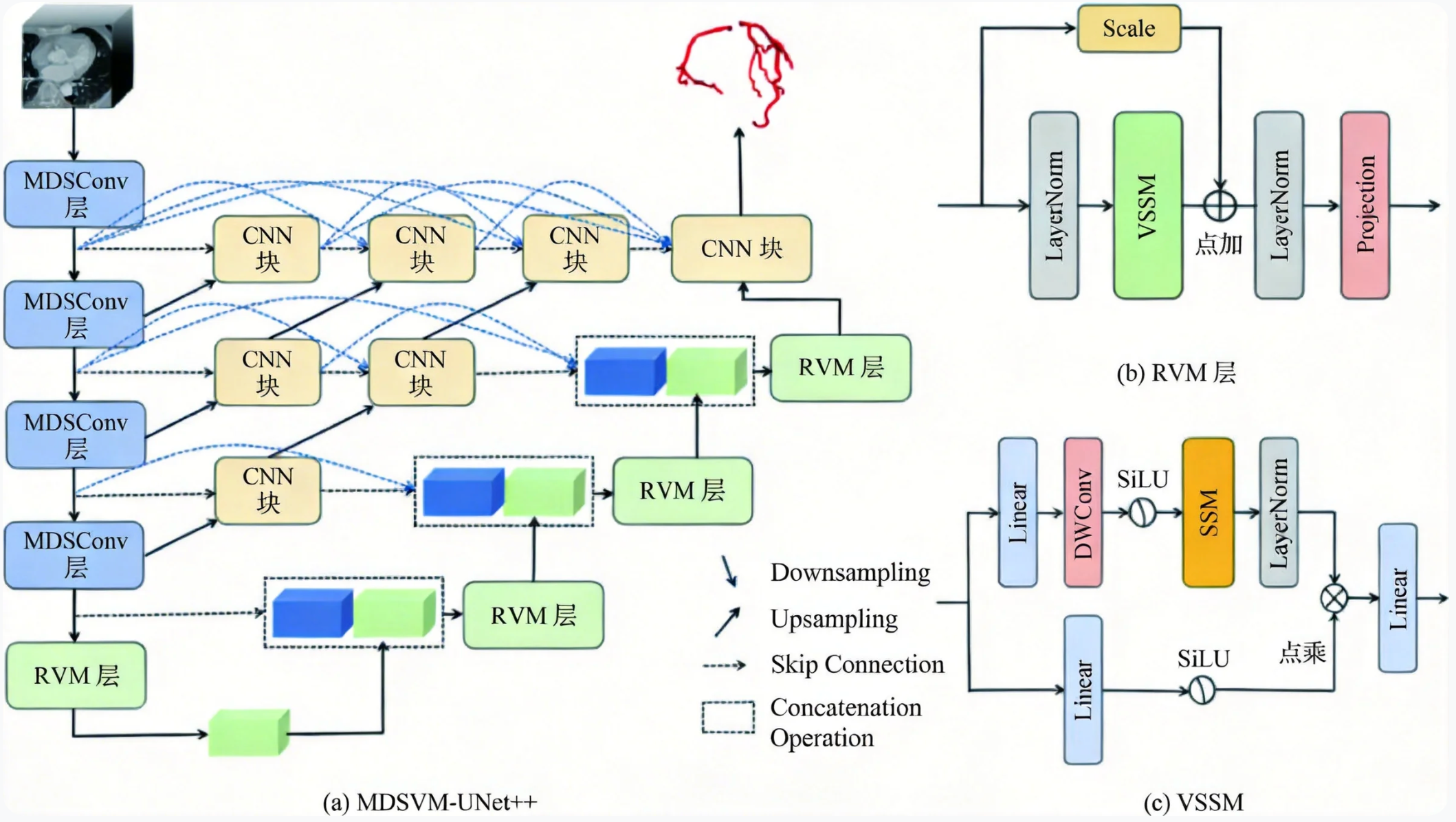}
    \caption{Detailed architecture of MDSVM-UNet. (a) The overall encoder-decoder network with UNet++-style dense skip connections. The encoder comprises four MDSConv blocks and one RVM block, while the decoder consists of three RVM blocks and a convolutional output layer. (b) The Residual Visual Mamba (RVM) layer with residual connection and scaling factor. (c) The Vision State Space Module (VSSM) with dual-branch parallel processing.}
    \label{fig:network}
\end{figure*}

As depicted in Figure~\ref{fig:network}(a), MDSVM-UNet adopts a hierarchical encoder-decoder architecture with UNet++-style dense skip connections. Given an input 3D medical image $\mathbf{I} \in \mathbb{R}^{C \times H \times W \times D}$, where $C$, $H$, $W$, and $D$ denote the number of channels, height, width, and depth (number of slices), respectively, the network is organized into five hierarchical levels with channel dimensions set to $[16, 32, 64, 128, 256]$.

\textbf{Encoder.} The encoder progressively extracts multi-scale feature representations through four MDSConv blocks followed by a single RVM block at the bottleneck. At each encoding level, the channel dimensionality is doubled while the spatial resolution is halved, enabling the network to capture both fine-grained local details and coarse-grained global semantics.

\textbf{Decoder.} The decoder comprises three RVM blocks and one $3 \times 3 \times 3$ convolutional layer for feature decoding and progressive spatial resolution recovery. At each decoding level, the channel dimensionality is halved while the spatial resolution is doubled via trilinear interpolation upsampling. The final transposed convolutional layer restores the output to the original input resolution.

\textbf{Skip Connections.} Following the UNet++ architecture \cite{zhou2020unetpp_redesign}, we employ dense skip connections that connect each decoder node to multiple encoder nodes at different hierarchical depths. Dense convolutional blocks within the skip pathways generate multi-scale intermediate representations, which are then channel-wise concatenated with decoder features to facilitate rich information flow and bridge the semantic gap between encoder and decoder representations.

\subsection{Multidirectional Snake Convolution}
\label{subsec:mdsconv}

\begin{figure*}[t]
    \centering
    \includegraphics[width=\columnwidth]{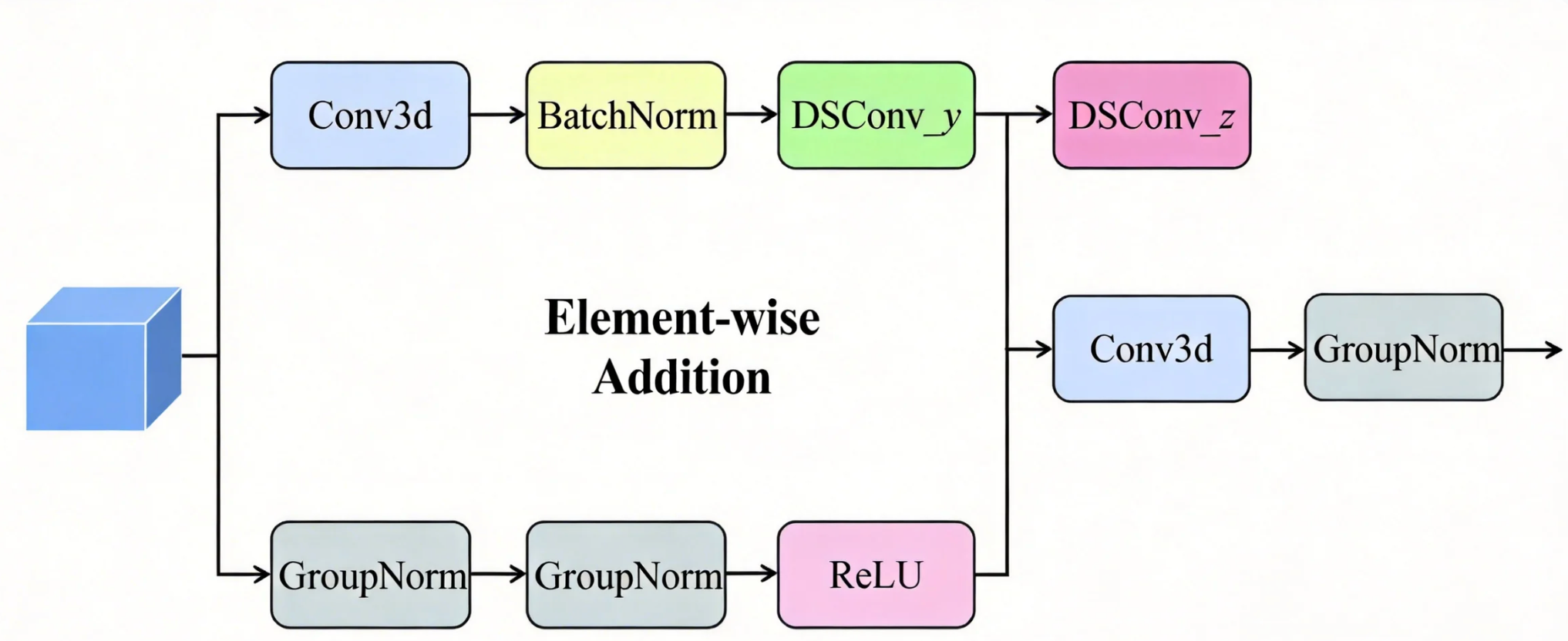}
    \caption{Architecture of the Multidirectional Snake Convolution (MDSConv) layer. Input features are processed by standard convolution and three axis-specific snake convolutions (along $x$, $y$, $z$ axes), followed by concatenation and fusion.}
    \label{fig:mdsconv}
\end{figure*}

\paragraph{Standard 3D Convolution Kernel.} Given a standard 3D convolution kernel with coordinates $\mathcal{K}$, the center coordinate is denoted as $K_i = (x_i, y_i, z_i)$. A $3 \times 3 \times 3$ kernel with unit dilation rate is represented as:
\begin{equation}
\mathcal{K} = \{(x-1, y-1, z-1), \ldots, (x+1, y+1, z+1)\}
\end{equation}

\paragraph{Snake Convolution with Iterative Offsets.} Following Qi et al.~\cite{qi2023dsconv}, we introduce deformable offsets $\Delta$ into the 3D convolution kernel, enabling the model to learn adaptive deformations while maintaining attention continuity along the tubular structure. An iterative propagation strategy is employed: starting from the center grid, offsets are progressively computed outward to prevent excessive receptive field dispersion. For each grid point $K_{i \pm c}$ in the kernel (where $c \in \{0, 1, 2, 3, 4\}$), the deformed coordinates along the $x$-axis are computed as:

\begin{figure*}[t]
\begin{equation}
\begin{cases}
(x_{i+c}, y_{i+c}, z_{i+c}) = (x_{i+c}, y_i + \sum_{i}^{i+c} \Delta y, z_i + \sum_{i}^{i+c} \Delta z) \\
(x_{i-c}, y_{i-c}, z_{i-c}) = (x_{i-c}, y_i + \sum_{i}^{i-c} \Delta y, z_i + \sum_{i}^{i-c} \Delta z)
\end{cases}
\label{eq:snake_x}
\end{equation}
\end{figure*}

Analogous iterative deformation strategies are applied along the $y$-axis and $z$-axis:
\begin{figure*}[t]
\begin{equation}
\begin{cases}
(x_{j+c}, y_{j+c}, z_{j+c}) = (x_j + \sum_{j}^{j+c} \Delta x, y_{j+c}, z_j + \sum_{j}^{j+c} \Delta z) \\
(x_{j-c}, y_{j-c}, z_{j-c}) = (x_j + \sum_{j}^{j-c} \Delta x, y_{j-c}, z_j + \sum_{j}^{j-c} \Delta z)
\end{cases}
\label{eq:snake_y}
\end{equation}
\begin{equation}
\begin{cases}
(x_{k+c}, y_{k+c}, z_{k+c}) = (x_k + \sum_{k}^{k+c} \Delta x, y_k + \sum_{k}^{k+c} \Delta y, z_{k+c}) \\
(x_{k-c}, y_{k-c}, z_{k-c}) = (x_k + \sum_{k}^{k-c} \Delta x, y_k + \sum_{k}^{k-c} \Delta y, z_{k-c})
\end{cases}
\label{eq:snake_z}
\end{equation}
\end{figure*}
\paragraph{Multi-View Anatomical Fusion.} Medical volumes inherently possess three orthogonal anatomical viewing planes---the sagittal, coronal, and axial planes---each providing complementary structural perspectives on the coronary vasculature. Multi-view feature fusion across these three planes enables more comprehensive observation of the target anatomy and facilitates the extraction of critical features hidden within complex coronary structures.

As illustrated in Figure~\ref{fig:mdsconv}, for a given input feature map $\mathbf{F}$, the MDSConv module computes four parallel feature representations: (1) a standard convolution extracts the baseline feature map $f^l(\mathbf{F})$; (2) three axis-specific snake convolutions extract directional feature maps $f^l(\mathbf{F}_x)$, $f^l(\mathbf{F}_y)$, and $f^l(\mathbf{F}_z)$ along the $x$, $y$, and $z$ axes, respectively. These four feature maps are concatenated channel-wise and fused through a subsequent convolution operation to produce the final feature map $\mathbf{X}^l$:
\begin{equation}
\mathbf{X}^l = f_{\text{ReLU}}\left(f_{\text{GN}}\left(\left[f^l(\mathbf{F}),\; f^l(\mathbf{F}_x),\; f^l(\mathbf{F}_y),\; f^l(\mathbf{F}_z)\right]\right)\right)
\label{eq:mdsconv}
\end{equation}
where $[\cdot]$ denotes channel-wise concatenation, GN refers to Group Normalization \cite{wu2018groupnorm}, and ReLU denotes the Rectified Linear Unit activation function. The fused multi-view features from the current layer are subsequently propagated to deeper layers for progressively refined feature extraction.

\subsection{Residual Visual Mamba Layer}
\label{subsec:rvm}

Given a feature map $\mathbf{F}^l \in \mathbb{R}^{C \times H \times W \times D}$, where $C$, $H$, $W$, and $D$ denote the channel count, height, width, and depth, respectively, the feature map is first reshaped and transposed into a sequence of shape $(L, C)$ where $L = H \times W \times D$. The RVM layer then captures global contextual information, after which the features are reshaped back to $(C \times 2, H, W, D)$.

\paragraph{Residual Connection with Scaling Factor.} As shown in Figure~\ref{fig:network}(b), the RVM layer first applies Layer Normalization (LN) to the input feature $\mathbf{x}^l_{\text{in}}$, followed by the Vision State Space Module (VSSM) for spatial long-range dependency modeling. A residual connection with a learnable scaling factor $\mathbf{s} \in \mathbb{R}^C$ is employed to enhance the modeling capacity of the SSM while introducing minimal additional parameters:
\begin{equation}
\tilde{\mathbf{Y}}^l = f_{\text{VSSM}}\left(f_{\text{LN}}(\mathbf{x}^l_{\text{in}})\right) + \mathbf{s} \times \mathbf{x}^l_{\text{in}}
\label{eq:rvm1}
\end{equation}

Subsequently, a second Layer Normalization and a linear projection transform $\tilde{\mathbf{Y}}^l$ into deeper feature representations:
\begin{equation}
\mathbf{Y}^l_{\text{out}} = f_{\text{Projection}}\left(f_{\text{LN}}\left(\tilde{\mathbf{Y}}^l\right)\right)
\label{eq:rvm2}
\end{equation}

\paragraph{Vision State Space Module (VSSM).} The VSSM takes the layer-normalized features $\mathbf{W}^l_{\text{in}} \in \mathbb{R}^{L \times C}$ as input and processes them through two parallel branches, as depicted in Figure~\ref{fig:network}(c).

In the first branch, the feature channels are expanded by a factor $\lambda$ through a linear layer, followed by depthwise separable convolution (DWConv), SiLU activation, SSM-based sequence modeling, and Layer Normalization:
\begin{equation}
\mathbf{W}_1 = f_{\text{LN}}\left(f_{\text{SSM}}\left(f_{\text{SiLU}}\left(f_{\text{DWConv}}\left(f_{\text{Linear}}(\mathbf{W}_{\text{in}})\right)\right)\right)\right)
\label{eq:vssm1}
\end{equation}

In the second branch, channel expansion and SiLU activation produce a gating signal:
\begin{equation}
\mathbf{W}_2 = f_{\text{SiLU}}\left(f_{\text{Linear}}(\mathbf{W}_{\text{in}})\right)
\label{eq:vssm2}
\end{equation}

The outputs of both branches are aggregated via Hadamard product and projected back to the original channel dimension $C$:
\begin{equation}
\mathbf{W}_{\text{out}} = f_{\text{Linear}}(\mathbf{W}_1 \odot \mathbf{W}_2)
\label{eq:vssm3}
\end{equation}
where $\odot$ denotes the element-wise Hadamard product. This dual-branch design enables the VSSM to selectively gate information flow, allowing the SSM branch to focus on modeling long-range spatial dependencies while the gating branch modulates the contribution of different feature channels.

\subsection{Decoder Architecture}
\label{subsec:decoder}

The decoder of MDSVM-UNet comprises three RVM blocks and one $3 \times 3 \times 3$ convolutional output layer. Given skip-connected features $\mathbf{F}^l \in \mathbb{R}^{C \times H \times W \times D}$ from the encoder and the output $\mathbf{P} \in \mathbb{R}^{C \times H \times W \times D}$ from the previous decoder block, feature fusion is first performed via element-wise addition, followed by RVM-based feature decoding and trilinear interpolation upsampling:
\begin{equation}
\tilde{\mathbf{P}} = f_{\text{Up}}\left(f_{\text{RVM}}(\mathbf{F}^l + \mathbf{P})\right)
\label{eq:decoder1}
\end{equation}
\begin{equation}
\mathbf{P}_{\text{out}} = f_{\text{Conv}}(\tilde{\mathbf{P}})
\label{eq:decoder2}
\end{equation}
where $f_{\text{Up}}$ denotes the trilinear interpolation upsampling operation and $f_{\text{Conv}}$ denotes the $3 \times 3 \times 3$ convolutional layer applied to the output of the final decoder block.

\subsection{Loss Function}
\label{subsec:loss}

Given the severe class imbalance between foreground coronary arteries and background tissue, the standard cross-entropy loss is unsuitable as it tends to bias the model toward the dominant background class. We instead adopt the Dice loss \cite{li2020dice}, which directly optimizes the Dice Similarity Coefficient and is well-suited for imbalanced segmentation tasks:
\begin{equation}
\text{Dice} = \frac{2 \times |A \cap B|}{|A| + |B|}
\label{eq:dice}
\end{equation}
\begin{equation}
\mathcal{L}_{\text{dice}} = 1 - \text{Dice}
\label{eq:diceloss}
\end{equation}
where $A$ and $B$ denote the sets of ground truth and predicted labels, respectively. Both stages of the segmentation pipeline employ Dice loss as the training objective.

% ============================================================================
% 4. EXPERIMENTS
% ============================================================================
\section{Experiments}
\label{sec:experiments}

\subsection{Dataset}
\label{subsec:dataset}

We evaluate the proposed method on the ImageCAS dataset \cite{zeng2023imagecas}, a large-scale public benchmark for coronary artery segmentation comprising 1,000 3D CTA images acquired from patients using a Siemens 128-slice dual-source CT scanner. During high-dose CTA reconstruction, the 30\%--40\% or 60\%--70\% cardiac phase was selected to obtain optimal coronary artery visualization. Each image has dimensions of $512 \times 512 \times (206\text{--}275)$ voxels, with in-plane resolution ranging from 0.29 to 0.43 mm$^2$ and inter-slice spacing of 0.25--0.45 mm. The left and right coronary arteries in each image were independently annotated by two radiologists, with cross-validation of annotations; in cases of disagreement, a third radiologist adjudicated, and the final labels were determined by consensus among all three annotators. Previous widely used coronary segmentation benchmarks \cite{kirisli2013evaluation, schaap2009evaluation} contain only 8 and 18 training images, respectively, making ImageCAS substantially more suitable for training and evaluating deep learning models.

\subsection{Implementation Details}
\label{subsec:implementation}

All experiments were implemented using Python 3.10 and PyTorch 2.1.1 with CUDA 11.8. Training and inference were conducted on a single NVIDIA GeForce RTX 3090 GPU with 24 GB of memory. We selected 750 CTA images from the ImageCAS dataset for training, with the remaining 250 images used for validation and testing.

\textbf{Stage 1 Training.} The first-stage segmentation network was trained for 25 epochs using the Adam optimizer with a learning rate of $1 \times 10^{-3}$.

\textbf{Stage 2 Training.} The second-stage segmentation network was trained for 50 epochs using the Adam optimizer, with learning rate decay by a factor of 0.1 applied at the 30th and 40th epochs.

\subsection{Evaluation Metrics}
\label{subsec:metrics}

We adopt three complementary metrics for comprehensive performance evaluation:

\begin{itemize}[leftmargin=*,itemsep=2pt]
    \item \textbf{Dice Similarity Coefficient (DSC):} Measures the volumetric overlap between predicted and ground truth segmentation masks, with values ranging from 0 (no overlap) to 1 (perfect overlap). Higher DSC indicates superior segmentation accuracy.
    
    \item \textbf{Hausdorff Distance (HD)} \cite{huttenlocher1993hausdorff}: Computes the maximum of the directed Hausdorff distances between the predicted and ground truth surface point sets:
    \begin{equation}
    H(A, B) = \max\left(h(A, B),\; h(B, A)\right)
    \end{equation}
    \begin{equation}
    h(A, B) = \max_{a \in A} \left\{\min_{b \in B} \|a - b\|\right\}
    \end{equation}
    where $A$ and $B$ represent the ground truth and predicted pixel sets, respectively.
    
    \item \textbf{Average Hausdorff Distance (AHD):} Provides a more robust boundary distance metric by averaging over all point-to-surface distances, reducing sensitivity to outliers compared to HD.
\end{itemize}
Lower HD and AHD values indicate better boundary alignment and spatial accuracy.

\subsection{Comparison with State-of-the-Art Methods}
\label{subsec:comparison}

\subsubsection{Single-Stage Segmentation Comparison}

We first evaluate the performance of our proposed segmentation network in the single-stage (Stage 1) setting, comparing against a comprehensive set of state-of-the-art coronary artery segmentation methods. Table~\ref{tab:stage1} presents the quantitative results. The ImageCAS baseline \cite{zeng2023imagecas}, which employs 3D U-Net in its first stage, serves as the primary benchmark.

\begin{table*}[t]
\centering
\caption{Quantitative comparison of single-stage segmentation methods on the ImageCAS dataset. Bold values indicate the best performance in each column. $\uparrow$ indicates higher is better; $\downarrow$ indicates lower is better.}
\label{tab:stage1}
\small
\begin{tabular}{lcccccc}
\toprule
\multirow{2}{*}{\textbf{Method}} & \multirow{2}{*}{\textbf{Params (M)}} & \multirow{2}{*}{\textbf{Loss}} & \textbf{DSC} $\uparrow$ & \textbf{HD} $\downarrow$ & \textbf{AHD} $\downarrow$ \\
\cmidrule(lr){4-6}
& & & & & \\
\midrule
FCN \cite{lei2020coronary} & 20.0 & $\mathcal{L}_{\text{dice}}$ & 0.5755 & 42.8031 & 1.6885 \\
 & & $\mathcal{L}_{\text{focal}}$ & 0.5573 & 37.0958 & 1.7619 \\
\midrule
3D U-Net \cite{cicek2016_3dunet} & 25.0 & $\mathcal{L}_{\text{dice}}$ & 0.6599 & 31.1518 & 1.1167 \\
 & & $\mathcal{L}_{\text{focal}}$ & 0.6521 & 30.9527 & 1.1966 \\
\midrule
TS ConvGRU \cite{kong2020tree} & 35.2 & $\mathcal{L}_{\text{dice}}$ & 0.6878 & 30.3350 & 1.4336 \\
 & & $\mathcal{L}_{\text{focal}}$ & 0.6766 & 32.4501 & 1.5864 \\
\midrule
GCN \cite{wolterink2019graph} & 41.5 & $\mathcal{L}_{\text{dice}}$ & 0.7061 & 27.8718 & 1.2433 \\
 & & $\mathcal{L}_{\text{focal}}$ & 0.6988 & 27.9970 & 1.4352 \\
\midrule
3D U-Net++ \cite{zhou2018unetpp} & 27.5 & $\mathcal{L}_{\text{dice}}$ & 0.6712 & 29.5178 & 1.0197 \\
 & & $\mathcal{L}_{\text{focal}}$ & 0.6710 & 28.8714 & 0.9840 \\
\midrule
DSU-Net \cite{qi2023dsconv} & 29.0 & $\mathcal{L}_{\text{dice}}$ & 0.6756 & 28.3684 & 0.9434 \\
 & & $\mathcal{L}_{\text{focal}}$ & 0.6643 & 35.4624 & 0.9711 \\
\midrule
LightM-UNet \cite{liao2024lightmunet} & 20.2 & $\mathcal{L}_{\text{dice}}$ & 0.6530 & 34.8181 & 1.0017 \\
 & & $\mathcal{L}_{\text{focal}}$ & 0.6446 & 46.4927 & 1.2864 \\
\midrule
UltraLight \cite{wu2024ultralight} & 0.238 & $\mathcal{L}_{\text{dice}}$ & 0.5745 & 50.9597 & 1.4438 \\
 & & $\mathcal{L}_{\text{focal}}$ & 0.5672 & 67.1512 & 1.3475 \\
\midrule
SwinUnet \cite{cao2022swinunet} & 256.3 & $\mathcal{L}_{\text{dice}}$ & 0.6723 & 31.8069 & 0.9270 \\
 & & $\mathcal{L}_{\text{focal}}$ & 0.6643 & 41.9588 & 1.2368 \\
\midrule
ImageCAS \cite{zeng2023imagecas} & 27.6 & $\mathcal{L}_{\text{dice}}$ & 0.6600 & 29.1486 & 0.9129 \\
\midrule
\textbf{MDSVM-UNet (Ours)} & \textbf{26.7} & $\mathcal{L}_{\text{dice}}$ & \textbf{0.6860} & 27.8430 & \textbf{0.9023} \\
 & & $\mathcal{L}_{\text{focal}}$ & 0.6814 & \textbf{27.5249} & 0.9260 \\
\bottomrule
\end{tabular}
\end{table*}

As shown in Table~\ref{tab:stage1}, our MDSVM-UNet achieves the highest DSC (0.686) and lowest AHD (0.9023) among all single-stage methods when trained with Dice loss. Notably, MDSVM-UNet surpasses the ImageCAS baseline by 2.61\% in DSC, 2.0032 in HD, and 0.2038 in AHD in the first-stage setting. Compared to the Transformer-based SwinUnet \cite{cao2022swinunet}, which requires 256.3M parameters, our method achieves competitive or superior performance with only 26.7M parameters---a reduction of approximately $9.6\times$ in model size. Furthermore, relative to DSU-Net \cite{qi2023dsconv}, which employs standard dynamic snake convolution, our MDSVM-UNet achieves an improvement of 1.04\% in DSC while reducing the parameter count by 2.3M, demonstrating the effectiveness of our multi-directional extension and the integration of visual Mamba for long-range dependency modeling.

\subsubsection{Two-Stage Segmentation Comparison}

Table~\ref{tab:stage2} presents the results of the two-stage segmentation pipeline, demonstrating the substantial benefits of our progressive coarse-to-fine strategy.

\begin{table*}[t]
\centering
\caption{Quantitative comparison of two-stage segmentation methods on the ImageCAS dataset. Bold values indicate the best performance in each column. ``--'' indicates experiments not conducted.}
\label{tab:stage2}
\small
\begin{tabular}{lcccccc}
\toprule
\multirow{2}{*}{\textbf{Method}} & \multirow{2}{*}{\textbf{Params (M)}} & \multirow{2}{*}{\textbf{Loss}} & \textbf{DSC} $\uparrow$ & \textbf{HD} $\downarrow$ & \textbf{AHD} $\downarrow$ \\
\cmidrule(lr){4-6}
& & & & & \\
\midrule
Two-Stage 3D U-Net \cite{cicek2016_3dunet} & 23.3 & $\mathcal{L}_{\text{dice}}$ & 0.7201 & 40.9693 & 3.0686 \\
 & & $\mathcal{L}_{\text{focal}}$ & 0.7183 & 32.7724 & 4.4687 \\
\midrule
Two-Stage DSU-Net \cite{qi2023dsconv} & 29.0 & $\mathcal{L}_{\text{dice}}$ & 0.7790 & 30.1524 & 0.9092 \\
 & & $\mathcal{L}_{\text{focal}}$ & 0.7720 & 36.4390 & 1.7171 \\
\midrule
Two-Stage LightM-UNet \cite{liao2024lightmunet} & 5.5 & $\mathcal{L}_{\text{dice}}$ & 0.8079 & 28.1864 & 0.9488 \\
 & & $\mathcal{L}_{\text{focal}}$ & 0.7848 & 28.5155 & 1.0054 \\
\midrule
Two-Stage SwinUnet \cite{cao2022swinunet} & 256.3 & $\mathcal{L}_{\text{dice}}$ & 0.8082 & 29.6208 & 1.0974 \\
 & & $\mathcal{L}_{\text{focal}}$ & 0.8012 & 28.2760 & 0.9625 \\
\midrule
Two-Stage FCN \cite{lei2020coronary} & -- & $\mathcal{L}_{\text{dice}}$ & 0.7950 & -- & -- \\
 & & $\mathcal{L}_{\text{focal}}$ & -- & -- & -- \\
\midrule
ImageCAS \cite{zeng2023imagecas} & 27.6 & $\mathcal{L}_{\text{dice}}$ & 0.7824 & 36.3886 & 1.7116 \\
 & & $\mathcal{L}_{\text{focal}}$ & 0.7748 & 36.1740 & 1.7261 \\
\midrule
\textbf{MDSVM-UNet (Ours)} & \textbf{26.7} & $\mathcal{L}_{\text{dice}}$ & \textbf{0.8365} & \textbf{27.8430} & \textbf{0.9023} \\
 & & $\mathcal{L}_{\text{focal}}$ & 0.8210 & 27.5249 & 0.9199 \\
\bottomrule
\end{tabular}
\end{table*}

Our two-stage MDSVM-UNet achieves a DSC of 0.8365, substantially outperforming the ImageCAS baseline (0.7824) by 5.41\%, while also achieving improvements of 8.5456 in HD and 0.8093 in AHD. These results confirm the effectiveness of our progressive coarse-to-fine segmentation strategy combined with the structure-aware encoding and efficient long-range decoding. Notably, the two-stage pipeline provides a significant performance boost over the single-stage variant: our DSC improves from 0.6860 (Stage~1) to 0.8365 (two-stage), representing a gain of 15.05\%, with corresponding improvements of 1.3056 in HD and 0.0106 in AHD. This substantial improvement underscores the complementary nature of whole-image coarse segmentation and block-level fine-grained refinement.

\subsection{Ablation Study}
\label{subsec:ablation}

To comprehensively analyze the contribution of each proposed component, we conduct ablation experiments on the ImageCAS dataset using the ImageCAS baseline as the reference model. Table~\ref{tab:ablation} summarizes the results.

\begin{table}[t]
\centering
\caption{Ablation study results on the ImageCAS dataset. Bold values indicate the best performance.}
\label{tab:ablation}
\small
\begin{tabular}{lccc}
\toprule
\textbf{Method} & \textbf{DSC} $\uparrow$ & \textbf{HD} $\downarrow$ & \textbf{AHD} $\downarrow$ \\
\midrule
Baseline & 0.7824 & 36.3886 & 1.7116 \\
+ MDSConv & 0.8241 & 27.5515 & 0.9046 \\
+ RVM & 0.8236 & 27.6082 & 0.9192 \\
+ MDSConv + RVM & \textbf{0.8365} & \textbf{27.8430} & \textbf{0.9023} \\
\bottomrule
\end{tabular}
\end{table}

\textbf{Effect of MDSConv.} Introducing multidirectional snake convolution alone improves DSC by 4.17\%, HD by 8.8371, and AHD by 0.8070 over the baseline, validating the effectiveness of multi-view deformable feature extraction for capturing the elongated tubular morphology of coronary arteries.

\textbf{Effect of RVM.} Incorporating the residual visual Mamba decoder alone yields improvements of 4.12\% in DSC, 8.7804 in HD, and 0.7924 in AHD, demonstrating the value of linear-complexity long-range dependency modeling for inter-slice feature propagation.

\textbf{Synergistic Effect.} When both MDSConv and RVM are combined, the model achieves the best overall performance with DSC of 0.8365 and AHD of 0.9023, surpassing the individual contributions of each module. This confirms that the two components are complementary: MDSConv provides structure-aware local feature extraction while RVM enables efficient global context modeling, together yielding a synergistic improvement in segmentation accuracy.

\subsection{Loss Function Comparison}
\label{subsec:loss_comparison}

We compare the effectiveness of two commonly used loss functions for addressing class imbalance: Dice loss and Focal loss \cite{lin2017focal}. As shown in Tables~\ref{tab:stage1} and \ref{tab:stage2}, Dice loss consistently outperforms Focal loss across all methods and evaluation metrics for coronary artery segmentation. This is attributed to the direct optimization of the Dice coefficient, which naturally accounts for the extreme foreground-background imbalance characteristic of coronary artery segmentation, where vascular structures occupy only a small fraction of the total image volume.

\subsection{Qualitative Results}
\label{subsec:visualization}

Figure~\ref{fig:visualization} presents a qualitative visualization of segmentation results from various methods alongside our proposed MDSVM-UNet. Visual inspection reveals that our method effectively reduces false positive artifacts compared to all baseline approaches, producing segmentation results with markedly improved vascular continuity and boundary smoothness. In particular, MDSVM-UNet demonstrates superior capability in preserving fine distal branches and maintaining topological connectivity of the coronary artery tree, which are clinically crucial for accurate stenosis detection and treatment planning.

\section{Conclusion}
\label{sec:conclusion}

In this paper, we presented MDSVM-UNet, a two-stage coronary artery segmentation framework that synergistically combines multidirectional snake convolution (MDSConv) and residual visual Mamba (RVM). MDSConv extends dynamic snake convolution to three orthogonal anatomical planes, enabling comprehensive multi-view capture of the elongated and tortuous vascular geometry. The RVM-based decoder efficiently models long-range inter-slice dependencies with linear complexity, significantly reducing parameters compared to Transformer-based alternatives. The proposed coarse-to-fine two-stage pipeline further reduces false positives while improving vascular continuity and smoothness. Experiments on the ImageCAS benchmark show that MDSVM-UNet outperforms the state-of-the-art by 5.41\% in DSC, 8.5456 in HD, and 0.8093 in AHD with only 26.7M parameters. Future work will focus on further model compression and validation on other vascular segmentation tasks and imaging modalities.

\clearpage
% ============================================================================
% REFERENCES
% ============================================================================
\bibliography{references}

\end{document}